\documentclass[aps,prb,preprint]{revtex4-1}
\usepackage{graphicx}
\usepackage{epsfig}
\usepackage{amsmath,amssymb}

\usepackage[breaklinks,colorlinks = true,linkcolor = blue,urlcolor=blue,citecolor=blue]{hyperref}

\begin{document}
\title{Random weights of DNNs and emergence of fixed points}

\author{L. Berlyand}
\affiliation{Department of Mathematics, Pennsylvania State University, University Park, Pennsylvania, 16802, USA}

\author{O. Krupchytskyi}
\affiliation{Department of Mathematics, Pennsylvania State University, University Park, Pennsylvania, 16802, USA\\\relax}

\author{V. Slavin}
\affiliation{B. Verkin Institute for Low Temperature Physics and Engineering of the National Academy of Sciences of Ukraine, Nauky Ave., 47, Kharkiv, 61103, Ukraine\\ email: slavin@ilt.kharkov.ua}

\begin{abstract}

This Letter is concerned with a special class of deep neural networks (DNNs) where the input and the output vectors have the same dimension. Such DNNs are widely used in applications, e.g., autoencoders. 
The training of such networks can be characterized by their fixed points (FPs). We are concerned with the dependence of the FPs number and their stability on the distribution of randomly initialized 
DNNs' weight matrices. Specifically, we consider the i.i.d. random weights with heavy and light-tail distributions. Our objectives are twofold. First, the dependence of FPs number and stability of 
FPs on the type of the distribution tail. Second, the dependence of the number of FPs on the DNNs' architecture. 
We perform extensive simulations and show that for light tails (e.g., Gaussian), which are typically used for initialization, a single stable FP exists for broad types of architectures. 
In contrast, for heavy tail distributions (e.g., Cauchy), which typically appear in trained DNNs, a number of FPs emerge. 
We further observe that these FPs are stable attractors and their basins of attraction partition the domain of input vectors. 
Finally, we observe an intriguing non-monotone dependence of the number of fixed points $Q(L)$ on the DNNs' depth $L$. 
The above results were first obtained for untrained DNNs with two types of distributions at initialization and then verified by considering DNNs in which the heavy tail distributions arise in training.

\end{abstract}

\maketitle

In recent  years, a variety of new technologies based on Deep Neural Networks (DNNs), also known as Artificial Neural networks (ANNs)
have been developed. AI-based technologies have been successfully used in physics, medicine, business and everyday life (see e.g., \cite{Yan-Yang:2022}).
The two key theoretical directions in DNN theory are the development of novel (i) types of DNNs and (ii) training algorithms.
 
One of the most important applications of DNNs is the processing of visual information \cite{Kaji-Kida:2019,Hong-Chen:2021}.
Image transformation (also known as image-to-image translation) means a transformation of the original image into another image according to the goals, for instance, enlarging the pictures without losing the quality. 
Another important example is the self-mapping transformation or autoencoder DNNs. Such DNNs are used e.g., for image restorations where the restored image is a fixed point (FP) of the DNN \cite{Mo-Wa-Zh:2022}.
Also, the proximity of a DNN's output vector to a FP can be used as a stopping criterion for DNNs' training. 

Note, that FPs of DNNs have many applications beyond image-to-image transformations.
In the modeling of the brain, fixed points appear in the time evolution of networks \cite{Fer-Mil:2000,Oz-Fin:2009,Rub-Hoo:2015,Ebsc-Ros:2018,Cur-Gen:2019}, whereas networks considered here are static.
Besides, the main part of the Firing model studies  deals with nonrandom weight matrices.
Another prominent example are Hopfield networks \cite{Hopfield:1982,Hopfield:1984}, where fixed points are used for memory modeling \cite{Krotov-Hopfield:2021}.
Hopfield model is also used in quantum physics, where FPs describe phase transitions \cite{Kimura-Kate:2024}.
Note, that in Hopfield networks the FPs of loss function are considered, while we study FPs of DNNs.

In this work, using numerical methods, we study the dependence of properties of fixed points on random distributions of i.i.d. weight matrices and on the network architecture.

\section{The model: Image-to-Image transformations and Fixed Points.}

We consider a fully-connected feedforward network where layer-to-layer transformation is a composition of the affine map with the nonlinear activation function \cite{Ber-Jab:2023}. The output vector $\boldsymbol x^{l+1}$ of the $l$-th layer of the DNN is
\begin{equation}
\boldsymbol{x}^{l+1}=\boldsymbol{\Phi}^{l}(\boldsymbol{x}^{l})=\varphi({\bf W}^{l}\boldsymbol{x}^{l}+{\bf b}^{l}),
\label{eq:one_layer}
\end{equation}
\noindent where $\boldsymbol{W}^l$ is a real-valued $n_{l+1}\times n_{l}$  weight matrix,
and $\boldsymbol{b}^l \in \mathbb R^{n_l}$ is a bias vector, the function $\varphi$ is the nonlinear activation function \cite{Ber-Jab:2023}. 
Then, DNN is a function $\boldsymbol \Phi$ that maps the input vector $\boldsymbol x^0$ into the output vector 
$\boldsymbol{x}^l$. 
\begin{equation}
\boldsymbol{\Phi}(\boldsymbol{x}^0)=(\boldsymbol{\Phi}^{L-1}\circ \boldsymbol{\Phi}^{L-2}\circ \cdots \circ \boldsymbol{\Phi}^{1}\circ \boldsymbol{\Phi}^{0})(\boldsymbol{x}^0)=\boldsymbol{x}^L.
\label{DNN_func}
\end{equation}

The fixed points are defined for autoencoder types of networks $\boldsymbol{\Phi}$, when the input and the output vectors have the same dimension, $n_0 = n_L$. Function $\boldsymbol{\Phi}$ is parametrized by weights and biases that hereafter will be denoted by $\alpha$, that is $\boldsymbol{\Phi} = \boldsymbol{\Phi}(\boldsymbol{x}, \alpha)$

Let us consider the problem of a single picture encoding and decoding \cite{Kin-Wel:2019,Wang-Cao:2022}. 
Let $\boldsymbol x_0$ corresponds to the original picture. For its encoding, we use a DNN $\boldsymbol{\Phi}_c$: $\mathbb{R}^{n_0} \rightarrow \mathbb{R}^{n_1}$:
$$\boldsymbol{\Phi}_c(\boldsymbol{x}^0)=\boldsymbol{x}^1,$$
\noindent where $\boldsymbol{x}^1 \in \mathbb{R}^{n_1}$ is the \emph{encoded picture} ($n_1$ is the size of  $\boldsymbol{x}^1$, in autoencoder DNNs \cite{Kin-Wel:2019},  $n_0>n_1$).
For  picture decoding we use another DNN $\boldsymbol{\Phi}_d: \mathbb{R}^{n_1} \rightarrow \mathbb{R}^{n_0}$. 
Let $\boldsymbol{\Phi}_d(\boldsymbol{x}^1)=\boldsymbol{x}^2$, where $\boldsymbol{x}^2\in \mathbb{R}^{n_0}$ is the  {\it decoded (restored) picture}.
Let DNN $\boldsymbol \Phi: \mathbb R^{n_0} \to \mathbb R^{n_0} $ be the composition:
$$\boldsymbol{\Phi}(\boldsymbol{x}^0) = (\boldsymbol{\Phi}_d \circ \boldsymbol{\Phi}_c)(\boldsymbol{x}^0) = \boldsymbol x^2.$$
 \noindent The goal of training of $\boldsymbol \Phi$ is to obtain the fixed point $\boldsymbol{x_2} = \boldsymbol{x_0}$:
$$\boldsymbol{\Phi}(\boldsymbol{x}^0) = \boldsymbol{x}^0.$$
This method of picture encoding/decoding can be used, for example, for employees access control. In this case employee's photo can be encoded using DNN $\boldsymbol{\Phi}_c$ and then securely transmitted via network 
to access server for decoding using $\boldsymbol{\Phi}_d$ and for access control.
Let $K$ be a number of employees. To perform training we start with input vectors $\boldsymbol{x}\in T_k$, $k=1,2,\ldots,K$. Here $T_k$ is the training set that contains  the photos of the  $k$-th employee. 
One of these photos, $\boldsymbol{x}^*_k \in T_k$, can be considered as  ``true'' photo of the employee stored on the access control server for identification. 
The other photos in $T_k$ are different  photos of the same employee (c.f., various fingerprints in a touch id, only one works).
Fixed points   $\boldsymbol{\Phi}(\boldsymbol{x}^*_k)=\boldsymbol{x}^*_k$ are obtained  via training  with the mean square  loss $L$ 
\begin{equation}
L(\alpha) = \sum_{k = 1}^K\sum_{\boldsymbol{x}\in T_k} 
\left\|\boldsymbol{\Phi}(\boldsymbol{x}, \alpha) - \boldsymbol{x}^*_k\right\|^2,
\label{loss-func1}
\end{equation}
c.f., fixed points in a special case of a single layer, deterministic, non-negative DNN in \cite{Pio-Cav:2024}.
Note that instead of \eqref{loss-func1} one can use,e.g., the cross-entropy loss function \cite{Bud:2017,Ber-Jab:2023}. There is no explicit formulas for $\boldsymbol\Phi_c$ and $\boldsymbol\Phi_d$ in this procedure. In order to restore a picture, one has to know all the weights obtained in training. This is a significant protection against hacking. 

\section{Light- vs Heavy-tailed distributions and DNN's training.}
We now explain how ``heavy-tailed'' distributions arise in DNNs.  Typical initialization of weights and biases is done with the light-tailed (subexponential) distributions, e.g. Gaussian. 
Such initializations are widely used for training via stochastic gradient descent (SGD) (see e.g, \cite{Bud:2017,Ba-Co:2020,Ca-Sc:2020,Gi-Co:2016,Li-Qi:2019,Ma-Co:2018,Ya:2020}). 
Note, that there are many modifications of SGD training based on random matrix theory (RMT) approaches aimed at improving DNNs' performance, e.g., Marchenko-Pastur pruning of singular values of random weight matrices enhances DNN's accuracy while reducing the noise \cite{Ma-Pa:1967,Be-Sa:2024}.
Numerical studies in the seminal work showed that the weigh matrices initialization by 
``light-tailed'' distribution {\it becomes ``heavy-tailed''} in the course of training.
This phenomenon is known as the  Heavy-Tailed Self-Regularization \cite{Ma-Ma:2021}. 
Moreover, recently it was shown that input-output Jacobian  of a trained DNN has also  ``heavy-tailed''  Empirical Spectral Distributions \cite{Pen-Sch:2023,Be-Sch:2024,Pa-Sl:2023,Ho-Ro:2019}.
Heavy-Tailed Self-Regularization allows us to use the tools of RMT for studying FP's properties of untrained and trained DNN's.

In the model of the pictures encoding/decoding,  a fixed point corresponds to a ``true'' photo of employee, $\boldsymbol{x}^*_k$, and the transition from ``lighty-tailed'' to ``heavy-tailed'' distribution during training will lead to drastic change in number of these FPs, their stability, and shapes/sizes of basins of attractions. 

\section{Fixed points and their basins of attraction.}
\label{Sec:FP}
Here we describe our numerical calculations of FP in untrained DNNs. 
For simplicity of the presentation  the dimension of input/output vectors $\boldsymbol{x}$ is taken $n=2$.
The space of input vectors $\boldsymbol{x}$ was chosen as a square: $\Omega = [-1,1]\times [-1,1] \subset \mathbb{R}^2$. 
This choice of $\Omega$ seems to be reasonable because the range of values of the main part of activation functions $\varphi$ is  $[-1,1]$.
This square was partitioned using grid with step $\delta= 0.05$, the grid points are:
\begin{equation}
\boldsymbol{x}_{j,l}=\left\{
\begin{array}{l}
x = -1 + \delta j, j=0,1,\ldots, \lfloor 2/\delta \rfloor \\
y = -1 + \delta l, l=0,1,\ldots, \lfloor 2/\delta \rfloor
\end{array}
\right. \,  ,
\label{x_jl}
\end{equation}
\noindent where $\lfloor\ldots\rfloor$ denotes an integer part. 
For each $\boldsymbol{x}_{j,l}$ we run  iterative process:
\begin{equation}
\boldsymbol{x}^{m+1} = \boldsymbol{\Phi}(\boldsymbol{x}^m), \quad m=1,2,3,\ldots,
\label{proc}
\end{equation}
where $\boldsymbol{x}^1 = \boldsymbol{x}_{j,l}$. 
For contraction mapping $\boldsymbol{\Phi}$ on a domain $\Omega$ Banach fixed-point theorem  guarantees convergence to a FP $\boldsymbol{x}^*$: 
\begin{equation}
\lim_{m\to\infty}\boldsymbol{x}^{m+1}=\boldsymbol{\Phi}(\boldsymbol{x}^m)= \boldsymbol{x}^*, \quad  \boldsymbol{x}^1 \in \Omega.   
\label{loop}
\end{equation}
The contraction property was checked numerically, and the existence of the limit \eqref{loop} was checked via Cauchy criteria:
$|\boldsymbol{x}^{m+1} - \boldsymbol{x}^{m}|<\varepsilon$, $m < N_0.$
In our calculations $\varepsilon =10^{-5}$, and $N_0=50$. If the limit exists, then $\boldsymbol{x}^{m}$ is the numerical approximation of fixed point 
$\boldsymbol{x}^*$ corresponding to starting grid point $\boldsymbol{x}^1=\boldsymbol{x}_{j,l}$ defined in \eqref{x_jl}
(for the details see Section \ref{Sec:NumProf}).

If domain $\Omega$ contains $Q>1$ fixed points and $Q$ basins of attraction 
$\Omega_k\in\Omega$, $k=1,2,\ldots, Q$, then for all grid points
$\boldsymbol{x}^1=\boldsymbol{x}
_{j,l}\in\Omega_k$ the limit \eqref{loop} provides a numerical approximation of the fixed point
$\boldsymbol{x}_k^*$.

We start from untrained DNN with depth $L=2$. Matrix entries and bias vector's components in \eqref{eq:one_layer}  are randomly initialized with normal distribution $N(0,\sigma_l)$, $\sigma_l = (n_{l})^{-1}$ $l=0,1$, 
where $n_{l}=\{2, 100\}$ are the layers widths (i.e., the weight matrices  sizes, $n_{l+1}\times n_l$, are: 
$100\times 2$ and $2\times100$).
Then unique fixed point $\boldsymbol{x}=0$ exists, i.e. $Q=1$, and the corresponding basin of attraction is the entire of $\Omega$.
This result can be interpreted as follows: such untrained DNNs can not identify ``true'' photos.

\begin{figure}[ht]
\begin{center}
\includegraphics[width=8.0cm]{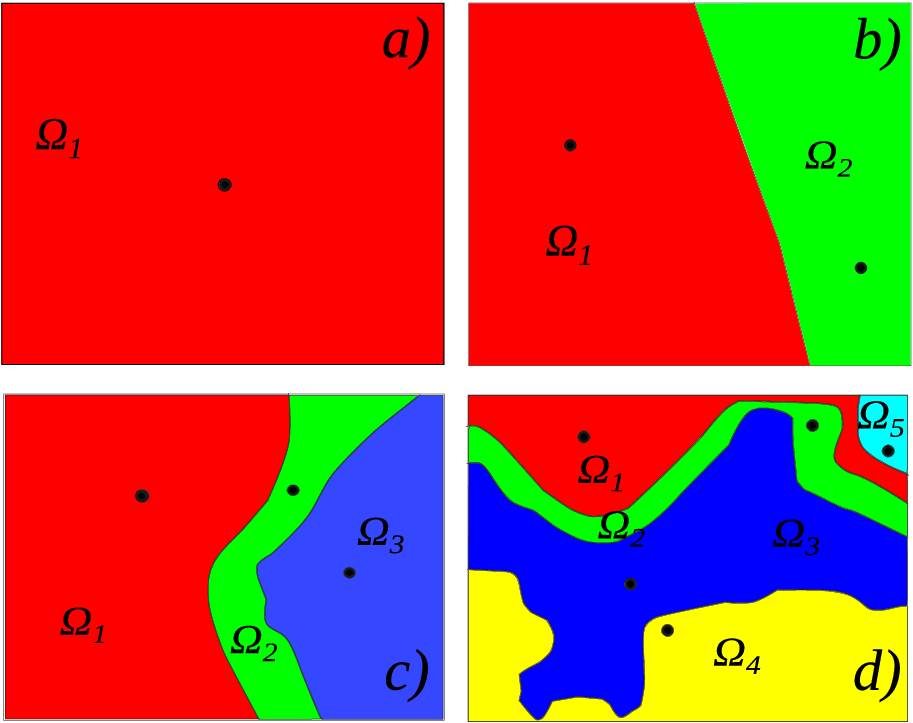}
\end{center}
\caption{(a) Normal distribution of the matrix elements and bias vector components. Number of layers $L=2$. The same result for Cauchy distribution and $L=20$.
(b) Cauchy distribution. $L=2$. (c) Cauchy distribution. $L=3$. (d) Cauchy distribution. $L=5$. The black circles are FPs. Different colors correspond to different basins of attraction $\Omega_k$.}
\label{fig_fp_basin}
\end{figure}

Next, using the approach based on Heavy-Tailed Self-Regularization (see e.g. \cite{Ma-Ma:2021,Pa-Sl:2023}) we model trained DNN by 
untrained DNN initialized by Cauchy  distribution centered at origin with scale $\gamma_l = (n_{l})^{-1}$.
The results are presented in Fig.~\ref{fig_fp_basin}b, c, and d.
Fig.~\ref{fig_fp_basin}b corresponds to the same architecture as in Fig.~\ref{fig_fp_basin}a ($L=2$, $n_{l}=\{2,100,2\}$), but we see two FPs, $Q=2$.
Fig.~\ref{fig_fp_basin}c corresponds to $L=3$, $n_l=\{2,100,100\}$, $Q=3$. 
In Fig.~\ref{fig_fp_basin}d  we present the results of calculations for
$L=5$, $n_l=\{2,100,\dots,100\}$, 
$Q=5$.
It is important, that {\it further increase in depth} $L$ leads to {\it decrease}  in $Q$ and the result for $L=20$ is the same as for $L=2$ and normal distribution --- the only one FP, $Q=1$. 
Due to the ``weak similarity'' effect \cite{Pa-Sl:2023} the choice of activation function $\varphi$ does not change the number of FPs.

Note, that the number of FPs, and the shapes/sizes of their basins of attraction $\Omega_k$ ($k=1,2,\ldots, Q$) are still random due to the finite size of the matrices. 
An interesting open question is the existence of deterministic limit of $Q$ and $\Omega_k$ as $n_l\to \infty$.

\begin{figure}[ht]
\begin{center}
\includegraphics[width=8.0cm]{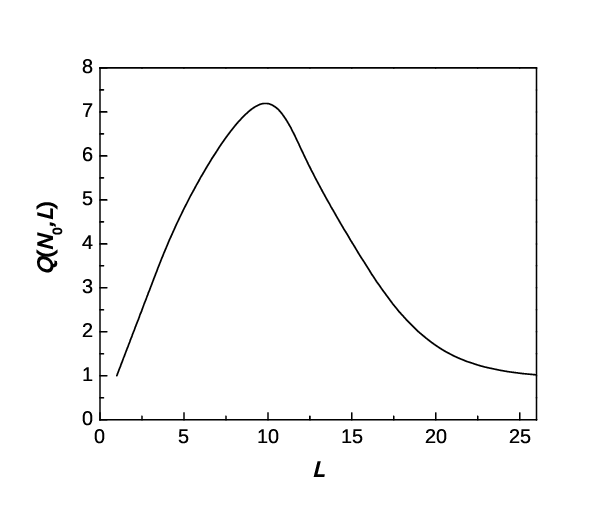}
\end{center}
\caption{Schematic illustration of the dependence of FP number $Q(N_0, L)$ on DNNs depth $L$ for DNNs with layers widths $N_0=100$ and with weights and biases initialized by random variables with Cauchy distribution.}
\label{fig_FP_L}
\end{figure}

Finally, we studied the number of FPs on DNN's depth $L$ and observed surprising non-monotone dependence of the most often appearing data (Mode) of $Q(N_0,L)$:
the number of FPs $Q$ first grows with DNN depth $L$, but then decreases.
The dependence of FP number $Q=Q(N_0,L)$ on  $L$ for DNNs with fixed layers widths $N_0=100$ and with weights and biases initialized via Cauchy distribution is presented schematically in Fig.~\ref{fig_FP_L}.

\section{Emergence of fixed points in trained DNNs.}
So far we studied the relation between randomness and FPs in untrained DNNs. 
However, training is the key ingredient in DNN's applications. Therefore in this section we address similar issues for trained DNNs and emphasize the similarities and differences.
The main difference is that the number of FPs in trained DNNs does depend on the DNN depths $L$ and it is determined by training set $T$ (e.g., it is the number of ``true'' photos 
in above example).

On the other hand training results in formation of a number of FPs similar to transition from light-tailed to heave-tailed 
distribution. The randomness in trained DNNs tends to vanish \cite{Sh-Je-Kru:2023} and direct application RMT is difficult \cite{Shch:2011,Russo:2020,Hi-Ke:20204}. However, the similarities between a trained DNN and untrained one via heavy-tailed self regularization allows us to use RMT tools for trained DNNs. Therefore the investigations of untrained DNNs with random initialization of the weight matrices and the investigations of trained DNNs complement each other.

\begin{figure}[ht]
\begin{center}
\includegraphics[width=8.0cm]{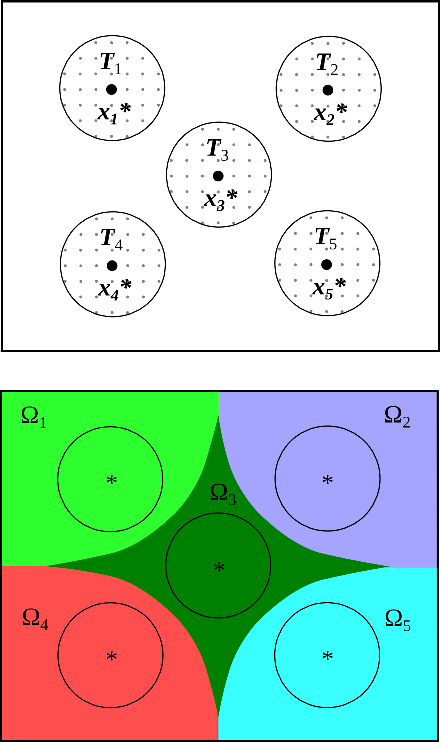}
\end{center}
\caption{The simplified model of employer's photo coding/decoding. 
Upper part corresponds to untrained DNN. Each employee's photo is represented by point in corresponding circle. The number of circles is the number of  
employers ($K=5$). Solid black circles  $\boldsymbol{x}^*_k$  correspond to the  ``true'' photos.
Lower part is the result of numerical calculation for trained DNN. The asterisk are the FPs.  We see numerically that the positions of FP ``*'' coincide with $\boldsymbol{x}^*_k$.
The filled areas are the corresponding basins of attraction $\Omega_k$.}
\label{fig6}
\end{figure}

We  use the same architecture for trained DNNs  as for untrained ones. The number of layers is $L = 3$, $n_l = \{2, 100, 100\}$, and activation function $\varphi(x)$ is ``HardTanh''.
The random initialization of DNN's parameters is done via Gauss distribution $N(0, \sigma)$, where $\sigma=1/\sqrt{L}$. 
DNN training is performed for a toy model of employee's photos coding/decoding. 
In this model the DNN's input/output photos are represented by 2-dimensional vectors ($\boldsymbol{x}^0, \boldsymbol{x}^L\subset \mathbb{R}^2$). 
It means that each employee's photo is represented by a point (upper part of Fig.~\ref{fig6}).
The photos of current employee are the points $\boldsymbol{x}$ inside the corresponding circle $T_k$, i.e., $\boldsymbol{x}\in T_k$.  These points form the training set for $k$-th employee ($k=1,2,\ldots, K$). 
The number of the circles is the number of employees, $K=5$. 
Centers of each circle are marked by solid black circle $\boldsymbol{x}^*_k$ and represent ``true'' photos.
The rest of points inside a circle represent ``non true'' photos of same employee. 
The loss function is chosen in the form  \eqref{loss-func1}. 

After training we search for FPs of the DNN. Similar to the case of untrained  DNNs we run the iterative process \eqref{proc} for each starting point $\boldsymbol{x}^1$ of the set $\boldsymbol{x}_{j,l}\subset\Omega$ (see \eqref{x_jl}).
If the process converges, then the corresponding FP is marked as $*$ (see lower part of Fig.~\ref{fig6}). 
Numerically we see that the positions of FPs coincide with ``true'' photos $\boldsymbol{x}^*_k$.
The different colors of subdomains 
in lower part of Fig.~\ref{fig6}  correspond to different FPs and their basins of attractions $\Omega_k$. 
In particular,  $T_k\subset \Omega_k$ that is 
each basin of attraction is larger then the corresponding training set. This means that this DNNs can identify photos outside training sets.

\section{Contraction mapping of DNN. Computer-assisted proof.}
\label{Sec:NumProf}
In  section \ref{Sec:FP} we show numerically that for the light-tailed distribution (Gauss distribution in our calculations) there exists a limit $\boldsymbol{x}^*$ of the process 
\eqref{proc} for all $\boldsymbol{x}^1 = \boldsymbol{x}_{j,l}\in \Omega$ (see \eqref{x_jl}) and $\boldsymbol{x}^* \to 0$ in the RMT limit.
Based on the existence of this limit, we concluded that $\boldsymbol{\Phi}$ provides a contraction mapping on $\Omega$, and therefore $\Omega$ is the basin of attraction with
$\boldsymbol{x}^* $ being the fixed point. Thus, we used implicitly the inverse Banach theorem, which is proved with a number of restrictions (see, e.g.\cite{Ja-Jo:20204}).
Here we prove explicitly (numerically) that for the light-tailed distribution of weight matrices $\boldsymbol{\Phi}$ provides a contraction mapping.  
Moreover, we study the influence of the DNN's number of layers $L$, the size of the weight matrices (the number of columns, $N$), the choice of activation function $\varphi$ and the variance of the distribution of the weight matrices $\sigma^2$. 

\begin{figure}[ht]
\begin{minipage}[ht]{0.49\textwidth}
\center{\includegraphics[width=1.0\linewidth]{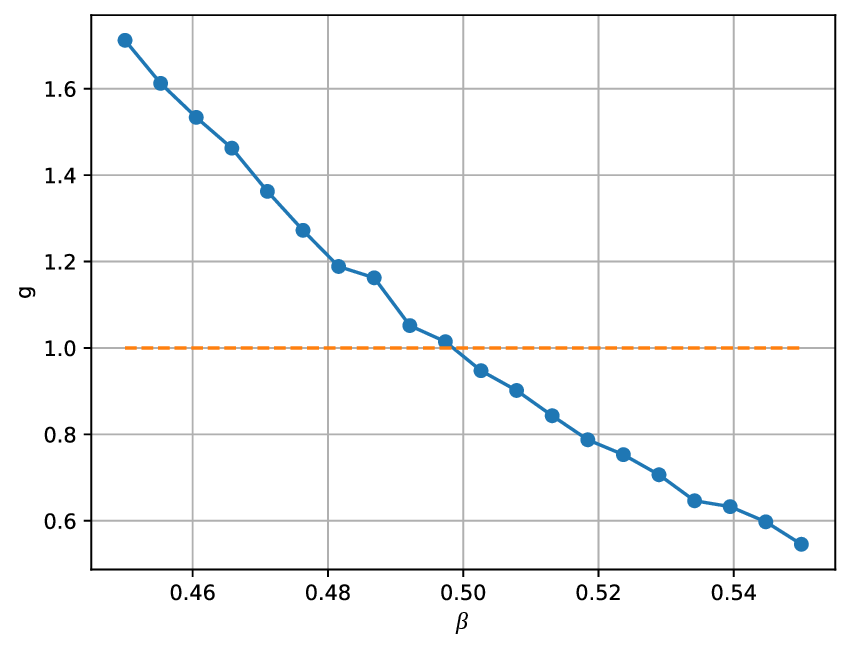}}
\end{minipage}
\begin{minipage}[ht]{0.49\textwidth}
\center{\includegraphics[width=1.0\linewidth]{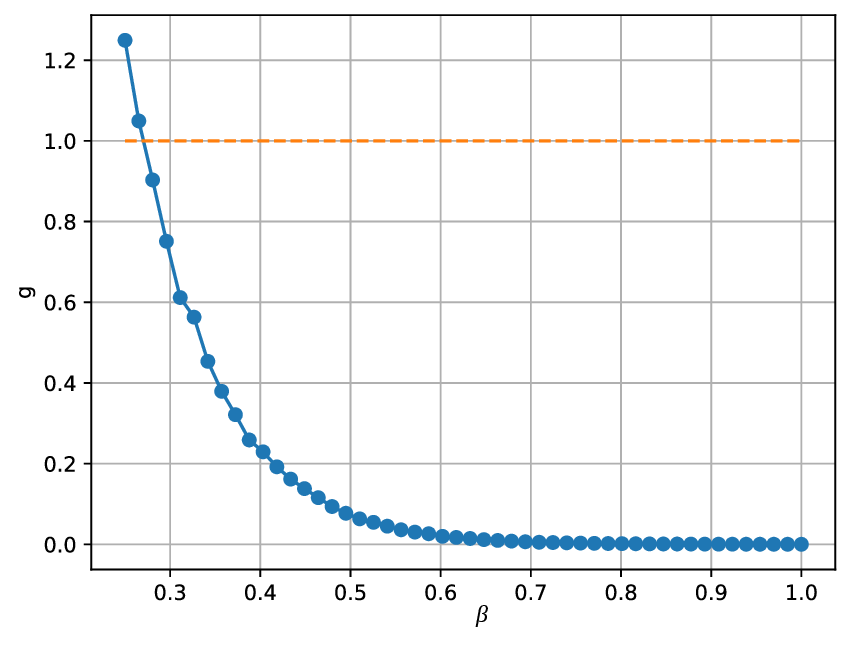}}
\end{minipage}
\caption{The dependence of contraction mapping parameter $g$ on $\beta$ for Gauss initialization of weight matrix. The  activation function is: 
a) ``Tanh''; b) `Sigmoid''. }
\label{fig:contraction_mapping}
\end{figure}

We choose $\sigma$ in the form
\begin{equation}
\sigma=N^{-\beta},
\label{variance_beta}
\end{equation}

\noindent and perform the calculations of  
\begin{equation}
\label{banch_criterion}
\end{equation}
$$g=\max_{\boldsymbol{x}_{j,l}\neq \boldsymbol{x}_{j',l'}} \frac{\lvert\boldsymbol{\Phi}(\boldsymbol{x}_{j,l})-\boldsymbol{\Phi}(\boldsymbol{x}_{j',l'})\rvert}{\lvert\boldsymbol{x}_{j,l}-\boldsymbol{x}_{j',l'}\rvert},$$ 

\noindent where grid points $\boldsymbol{x}_{j,l}$  are defined in \eqref{x_jl}. 
The dependence of $g$  defined in \eqref{banch_criterion} on the parameter $\beta$ \eqref{variance_beta} for a single layer with Gauss initialization of the
weight matrix ${\bf W}$, size $N=400$, and ``Tanh'' activation function is presented in Fig.~\ref{fig:contraction_mapping} a). 
We see that the existence of the contraction mapping (i.e., $g<1$) depends crucially on $\beta$.  There exists a critical value $\beta_{cr}\approx1/2$, so that
for $\beta \geq \beta_{cr}$ the function $\boldsymbol{\Phi}$ provides contraction mapping on $\Omega$.

Indeed, let us rewrite \eqref{eq:one_layer} as the following:
$$\boldsymbol{x}^{l+1} = \varphi(\boldsymbol{y}^l), \quad  \boldsymbol{y}^l={\bf W}^l\boldsymbol{x}^l + {\bf b}^l.$$ 

For light-tailed initialization of ${\bf W}$ and ${\bf b}$ using the Central Limit Theorem (i.e., in the limit of $N\to \infty$) we obtain that
each component of $\boldsymbol{y}^l$ is a random variable with variance $\Sigma^2=\text{Var}[\left(\boldsymbol{y}^l\right)_k] \sim (N+1)\sigma^2$.
Note that for a stable distribution, like Gaussian this statement is true and true for limited $N$. 
Thus, from the requirement of boundedness of $\Sigma^2$ it follows that $\sigma\sim N^{-1/2}$. i.e., $\beta_{cr} = 1/2$ (see. e.g., \cite{Pen-Sch:2023,Pa-Sl:2023}).

We assume that {\it odd activation functions} (like ``Tanh'', ``HardTanh'') {\it do not affect} $\beta_{cr}$, since they
have little effect on the tails of the distribution and provide symmetric limiting of the central peak.
Our numerical calculations for different odd activation functions confirm this assumption. 

At the same time, for the ``sigmoid'' activation function 
$$\varphi(x) = \frac{1}{1 + \exp(-x)},$$

\noindent (which is neither even nor odd) the critical  $\beta_{cr}$ exists with $\beta_{cr} \approx 1/4$ (see Fig.~\ref{fig:contraction_mapping} b)).

\begin{figure}[ht]
\begin{center}
\includegraphics[width=8.0cm]{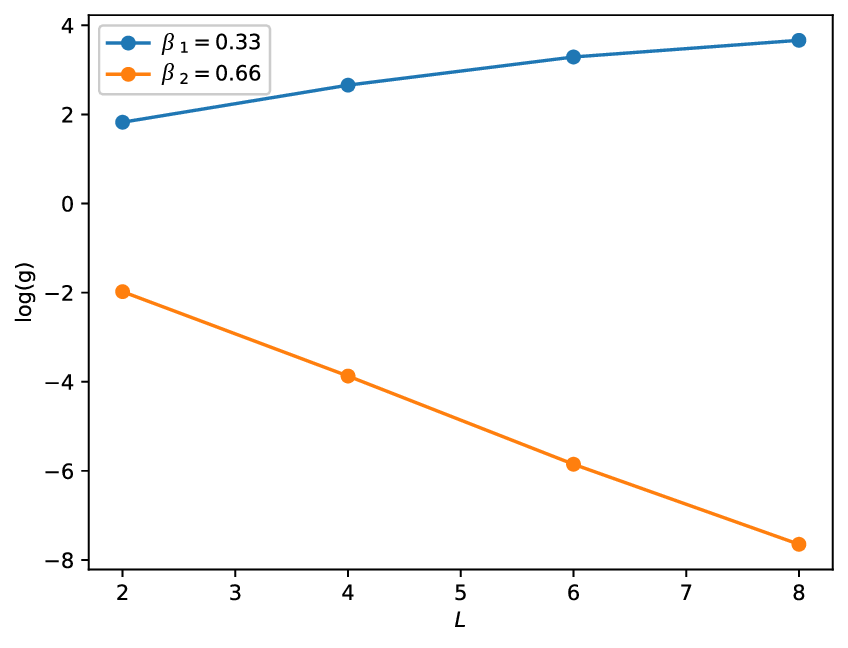}
\end{center}
\caption{The dependence of the contraction mapping parameter $g$ on the number of DNN's layers $L$ for different values of $\beta_1 < \beta_{cr}$ and  $\beta_2> \beta_{cr}$. }
\label{fig:contraction_mapping_L}
\end{figure}

The dependence of the contraction mapping parameter $g$ on the number of layers $L$ is presented in Fig.~\ref{fig:contraction_mapping_L}).
Linear dependence in semi log-scale reflects the composition structure of DNN (see \eqref{DNN_func}): let  $g_0$ be a contraction mapping parameter of single layer, then

\begin{equation}
g=g_0^L, 
\label{g_L}
\end{equation}

\noindent as it seen in Fig.~\ref{fig:contraction_mapping_L}.

Our numerical calculations for heavy-tailed distributions (Cauchy distribution in our calculation) also show full correspondence between the areas of 
contraction of $\boldsymbol{\Phi}$ (i.e. the areas where  $g<1$) and basins of attraction of $\boldsymbol{\Phi}$.

Thus, our numerical calculations show:
\begin{itemize}
\item The existence of contraction mapping of DNN depends crucially on the parameter $\beta$ in \eqref{variance_beta}.
There exists critical value of  $\beta= \beta_{cr}$ which separates the areas of contraction and non-contraction mappings.
\item The width of weight matrices $N$ only affect on the normalization of variance $\sigma^2$ (see \eqref{variance_beta}).
\item We see some sort of universality of the obtained results with respect to the activation function $\varphi$ choice: odd activation functions do not affect $\beta_{cr}$.
\item The dependence of the mapping parameter $g$ on the number of layers $L$ is exponential \eqref{g_L}. This result reflects the composition structure of DNNs.
\end{itemize}

\section{Summary of Results and discussion}
We studied the relation between random distributions of weights and the properties of fixed points in autoencoder DNNs.
We first considered {\it untrained} DNNs with random initialization of weight matrices according to {\it ``light-tailed''} distribution, e.g., Gaussian.

We showed the existence of the unique fixed point (FP) for such Gaussian DNN's with arbitrary layer width and depth, and for a wide class of ``sigmoid'' - type activation functions. 
In context of image encoding/decoding problem it means that these DNNs can not identify ``true''  photos.

We next considered {\it untrained} DNNs with random
{\it ``heavy-tailed''} distribution (e.g., Cauchy)
initialization of weight matrices.
Here a number of FPs appeared 
in contrast  with the  ``light-tailed'' case.
We further established the stability of these FPs 
and determined their basins of attractions. 
Our study showed surprising non-monotone dependence of the number of DNN's fixed points, $Q(N_0,L)$, 
on DNN depth $L$ (Fig.  \ref{fig6}).

We finally considered  {\it trained}  DNNs and showed that training with standard 
``light-tailed'' initialization  leads to formation of many stable FPs analogous to untrained ``heavy-tailed'' DNNs.
In image encoding/decoding problem the number of ``true'' photos is prescribed. Therefore the goal of DNN's training is to obtain the same number $Q$ of FPs by choosing optimal DNN's architecture, which can be done using, e.g., $Q(N_0,L)$ dependence
obtained above.
Moreover, we showed that each basin of attraction,
$\Omega_k$, $k=1,2,\ldots$, contains the corresponding training set  $T_k\subset \Omega_k$ (Fig.~\ref{fig6}). 
Thus, the DNNs can identify photos outside training set, 
which can be used, e.g. for cross-validation of training.

\section{Acknowledgments}
The work of V.S. was supported by Grant ``International Multilateral Partnerships for Resilient Education and Science System in Ukraine'' IMPRESS-U: N7114
funded by US National Academy of Science and Office of Naval Research. 
The work of L.B. was partially supported by the NSF Grant IMPRESS-U: N2401227.
The authors are grateful to Dr. Mikhail Genkin, Dr. Vladimir Itskov, and Dr. Ievgenii Afanasiev for many discussions and useful suggestions.
The authors are also grateful to Zelong Li for the participation in numerical examples in the Section ``Emergence of fixed points in trained DNNs'' and for the help in creating  Figure 3.

\end{document}